**Overcoming technical adoption barriers for mobile service robots in rehabilitation**

Christian Sternitzke[1], Sebastian Blumenthal[1], Lukas Kleedörfer[2], Verena Deserno[1], Anke Mayfarth[1]

[1] TEDIRO Healthcare Robotics, Ehrenbergstr. 11, 98693 Ilmenau, Germany

[2] MetraLabs GmbH Neue Technologien und Systeme, Weimarer Str. 28, 98693 Ilmenau, Germany

**Abstract**

Many publications on robotic systems in healthcare describe early-stage work on low technology readiness levels. This paper describes how a mobile service robot approved as a medical device reaches higher technology readiness levels by adding peripheral functions and smaller improvements, which are pivotal for user acceptance in clinical environments and which often cannot be elicited by questioning users' ex-ante as certain aspects only come in mind from testing the systems in clinical settings/operational environments. Especially developers of service robots in healthcare are advised to plan with such downstream developments, which can take significant implementation time, to obtain user acceptance and achieve widespread adoption of their robotic systems.

## 1 Introduction

The aging population in many countries poses two important challenges: increasing costs in the healthcare systems by older and, consequently, more frail patients, and an increasing gap of skilled labor in care work. Service robots are seen to tackle these challenges. The space of possibilities how service robots could help is rather broad, covering intelligent robots for medical functionality and physiological monitoring, autonomous assistive tasks robots like for providing food, rehabilitation robots, or companion robots for social interaction Vercelli et al. (2017). Most of the robots described in the literature are research robots showing the proof of principle of basic functionalities on low technology readiness levels.

This work concentrates on service robots in the wider rehabilitation space that mobilize people by offering different walking exercises. The present dominant designs for commercially available rehabilitation robots with focus on the lower limbs are either exoskeletons or treadmill systems like the Lokomat (Jezernik et al., 2003). Both have in common the fact that they mostly address people with severe walking impairments that, at the same time, need assistance for using such robotic systems. Contrasting this approach, the service robot system described below addresses patients in late-stage rehabilitation and elderly care which can walk alone, but that nevertheless benefit from feedback during their walking exercises, let it be for increasing their motivation and possibly adherence for the training, or providing explicit feedback regarding walking patterns. More specifically, the work presented here pictures the THERY robot platform, a mobile service robot with two walking exercises applications: THERY UAG that provides feedback for proper crutch use and walking patterns for patients using forearm supports, and THERY GO that offers walking exercises which are independent from the assistive device of the robot user. Both versions are commercially available, and it is described how improvements were made from lower technology readiness levels of THERY UAG to higher ones to reach a stage that is commercially relevant. Such late-stage development efforts are pivotal for having a system in place that is sufficiently robust, a requirement needed so that users have enough trust for the actual deployment (Lu et al., 2020). But it is more than robustness, it is about features that assure smooth robot operations beyond the core clinical functionality. Such implementations pose a further challenge for robot manufacturers when deploying robotic systems in healthcare facilities.

The remainder of this article is as follows: Section 2 provides insights into the clinical motivation behind the endeavor. Section 3 gives more details on service robots for walking trainings. Section 4 explains the THERY platform with the THERY UAG application in more detail, while section 5 describes how field tests helped reshape product requirements on peripheral functions to increase the technology readiness level of the whole system. The last section summarizes the results.

## 2 Clinical motivation for walking training

Walking capacity is a decisive factor for independence, functional autonomy, and quality of life across older adults and diverse patient groups. In senior homes, rehabilitation centers, and general hospitals, residents and patients frequently present with frailty, mobility limitation, neurological impairment, or reduced endurance, leading to increased care needs and a higher risk of falls and secondary complications. Evidence-based practice therefore supports targeted walking training as a core intervention domain rather than an adjunctive or optional activity (Sherrington et al., 2020), and high-level evidence indicates that walking-based interventions can improve mobility-related outcomes, particularly in frail older adults, in patients after stroke and in patients undergoing orthopaedics surgery.

In orthopaedics, treadmill training after hip surgery offered at a higher intensity than conventional gait training significantly improves functional outcomes, gait symmetry, hip range-of-motion, or abductor strength (Hesse et al., 2003). Walking skill training programs proved to increase various functional outcomes like walking duration (in terms of the 6 Minute Walk Text) or step climbing after hip surgery (Heiberg et al., 2012). Step count and walking time have a positive effect on functional scores from ankle and tibia fracture patients (North et al., 2023), implying that walking training benefits patients after multiple forms of lower limp surgery.

In community-dwelling older adults, structured exercise programs reduce falls, and walking-oriented training is recognized as a practical and safe way to promote functional mobility, even though walking-only interventions show less certain evidence than balance- and strength-oriented multimodal programs. In frail or prefrail older adults, structured walking protocols have demonstrated favorable effects on physical function, frailty indicators, and endurance, supporting the feasibility of walking practice in residential care settings. These findings justify walking training as a scalable and broadly applicable intervention in geriatric and long-term care environments (Ishigaki et al., 2024; Karttunen et al., 2015; Wu et al., 2024).

In stroke rehabilitation, the evidence base for walking training is particularly strong. Systematic reviews and meta-analyses of randomized controlled trials show that specific treadmill and overground walking training improves walking speed, walking distance, balance, motor function, participation, and, to some extent, self-care. These benefits are typically associated with repeated, task-specific practice over several weeks, emphasizing that walking training is not merely a compensatory strategy but a core therapeutic target. This evidence provides a solid foundation for integrating structured, robot-assisted walking training into rehabilitation workflows (Collins et al., 2018; Rubin et al., 2025; Wu et al., 2024).

To conclude, walking-focused interventions benefit older adults and several patient groups, especially in frailty, fall-risk and stroke-related mobility disorders. By extending these principles into senior homes, rehabilitation centers, and clinical settings, robot-assisted walking exercises can improve functional outcomes, enhance training safety, and standardize assessment across institutions (Ishigaki et al., 2024; Karttunen et al., 2015; Wu et al., 2024).

## 3 Service robots for late-stage walking training

For tackling labor shortages in healthcare, service robots must provide a high level of autonomy, and ideally operate in a self-service mode. This contrasts established rehabilitation robot systems (for reviews, see e.g. Banyai and Brișan (2024); Mohebbi (2020)) like exoskeletons or treadmill systems. These are typically used by patients that cannot walk without assistance and that need support in utilizing those rehabilitation robot systems, like being affixed to an exoskeleton or treadmill-based robot like the Lokomat, mainly addressing neurological conditions.

In contrast, Gross, Scheidig, et al. (2017) describe a service robot developed in the ROREAS project offering walking and orientation training for stroke patients as self-training, requiring no therapist-in-the-loop for the exercises performed by the human-robot tandem. The robot in the ROREAS project was tested in field trials with patients in a German rehabilitation hospital (technology readiness levels 5/6). During the training sessions, the robot used its display and voice outputs for providing guidance where to go, and it followed the patients during their walking exercises. Gross, Meyer, et al. (2017), in a companion publication, describe that two-thirds of the patients during the field trials showed a higher training motivation when being accompanied by the robot. Some of them reported that they covered longer walking distances. Somewhat related to that work are two experimental field studies from Japan where a mobile service robot was used in a field trial to accompany elderly. The first trial involved mainly men in their late sixties, and the second trial predominantly women in their late eighties from a nursing home, many also suffering from dementia. They walked with and without the robot. Being questioned afterwards, the probands stated that they preferred walking with the robot rather than alone (Karunarathne et al., 2019; Nomura et al., 2021).

While these robot walking applications are rather simplistic, more complex approaches have been described in the literature as well, such as the robot developed in the ROGER project (Röhner et al., 2021; Scheidig et al., 2021) that offered gait training for patients following hip replacement surgery (technology readiness level 6). Here, a mobile service robot was used that captured patients during crutch walking with its 3D camera, analyzed the walking patterns of the patients and their crutch motions using motion capture technologies, and provided audiovisual corrective feedback. For this system, it was also reported that patients using it were highly motivated (Meyer et al., 2026). By using a robot of this kind, therapists can delegate repetitive tasks to the robot and focus on more complex hands-on therapies – an area that is often neglected in times of staff shortages.

These results regarding patient motivation indicate that service robots may perform an important function in healthcare: they may increase adherence to training plans and, ultimately, clinical outcomes. However, robust evidence for this relationship is still lacking.

Besides the work outlined above, Guffanti and colleagues describe a mobile robotic platform with a similar configuration as in the ROGER project for gait analysis, where a 3D camera and motion capture software was employed to elicit spatiotemporal and kinematic gait patterns of a person. This early technology demonstrator has partially been tested in clinical settings already (Guffanti et al., 2024; Guffanti et al., 2021). Somewhat related to this work is the mobile service robot developed for stroke patients and described in Lee et al. (2022). It projected recommended foot positions on the ground during training and made use of LIDARs to track the actual foot positions. This system was validated in the lab.

Overall, the literature only describes fully qualified robotic training systems for neurology like the Lokomat (Jezernik et al., 2003), while there is a gap in the literature for solutions addressing orthopedic or geriatric applications, in particular with mobile service robots. As the interaction level of autonomous mobile service robots with their environment with its dynamic and stationary obstacles, getting in touch with various user groups like patients, therapists, and nurses is much higher than that of stationary systems. Consequently, stakeholder requirements of these user groups must be met so that these autonomous systems - in order to be accepted - can reach high technology readiness levels. As the different stakeholders cannot articulate their requirements from pure imagination, field tests with them are necessary in actual operational environments, and it can be anticipated that these requirements address many facets. Below, it is described which functionalities were implemented to reach higher technology readiness levels.

## 4 The THERY platform and its functions

The robot prototype of the ROGER project was subsequently further developed into a medical device-grade mobile robot called THERY, and together with the crutch training application named THERY UAG, it was certified as a medical device within the European Union. The system architecture of the prototype was already laid out in (Scheidig et al., 2021). Consequently, only important building blocks are subsequently described, such as the hardware and software configuration, different functionalities, and its first application UAG for gait training on forearm supports

## 4.1 Hardware configuration

### *4.1.1 Basic hardware configuration*

The THERY robot uses a mobile base provided by MetraLabs GmbH (Germany) and developed for their TORY shelf-scanning robots, but with a 60Ah battery to extend the robots' operating time beyond 8 h. The robot is powered by a differential drive, allowing a speed of up to 0.8 m/s. Application software runs on a high-performance PC with a Nvidia GPU capable for motion capture algorithms.

The footprint of the 60 kg robot platform is 50 cm in diameter. This means that the THERY robot can rotate on the spot without the structure hitting an obstacle. With a height of 1.5 m, THERY is a bit smaller than most people training with it.

### *4.1.2 On-board safety sensors*

The robot uses four different sensors for safe navigation: First, a SICK LIDAR with safety fields senses the environment in driving direction about 10 cm above the ground and is also used for mapping. It is complemented by an obstacle avoidance stereo camera located on the upper part of the robot to discover obstacles not sensed by the LIDAR. A safety strip/bumper surrounds the robot housing and stops the robot upon in contact with an obstacle. Functional safety is assured by a control unit achieving ISO 13849-1 performance level d and works independently of the application software layer. This control unit reduces the robots' speed to a safety speed upon detection of obstacles within the safety zones of the LIDAR.

### *4.1.3 Human-robot interaction*

For contactless human-robot interaction, a 3D camera is used (Kinect Azure), and user login on the robot is realized via an RFID transponder that is read out by the robot (for details, see Section 4.4 below). There is an upright 17-inch touch-display that offers display-based user interaction. Outputs are pre-defined and shown on the display as well as given via speech outputs.

### *4.1.4 Robot design*

The design of THERY was chosen to stimulate effects of anthropomorphism (Bartneck et al., 2009; Epley, 2018; Złotowski et al., 2015) [1], a phenomenon for which a meta-analysis has found that it has a significant and positive effect on the intention to use a robot and on the robots' perceived intelligence (Blut et al., 2021). The effect is also present for training activities like weight-loss coaching, as experiments could show (Kidd and Breazeal, 2008). In the study byScheidig et al. (2021), two robotic platforms with different levels of embodiment features were compared. The platform with the more pronounced embodiment features received higher user ratings, as measured by the System Usability Scale (SUS) score. Based on these findings, THERY was deliberately designed with a more pronounced figure-shape

[1] "*Anthropomorphism refers to the attribution of a human form, human characteristics, or human behavior to nonhuman things such as robots, computers, and animals*" (Bartneck et al. 2009, p. 74). Złotowski et al. (2015) emphasize that anthropomorphism is more than appearance, also including the actual way of user interaction. Epley (2018) similarly defines anthropomorphism as the "*attribution of humanlike traits*".

(see Fig. 1) and uses an animated face in the upper part of the display above the output section (see Figs. 3 and 4).

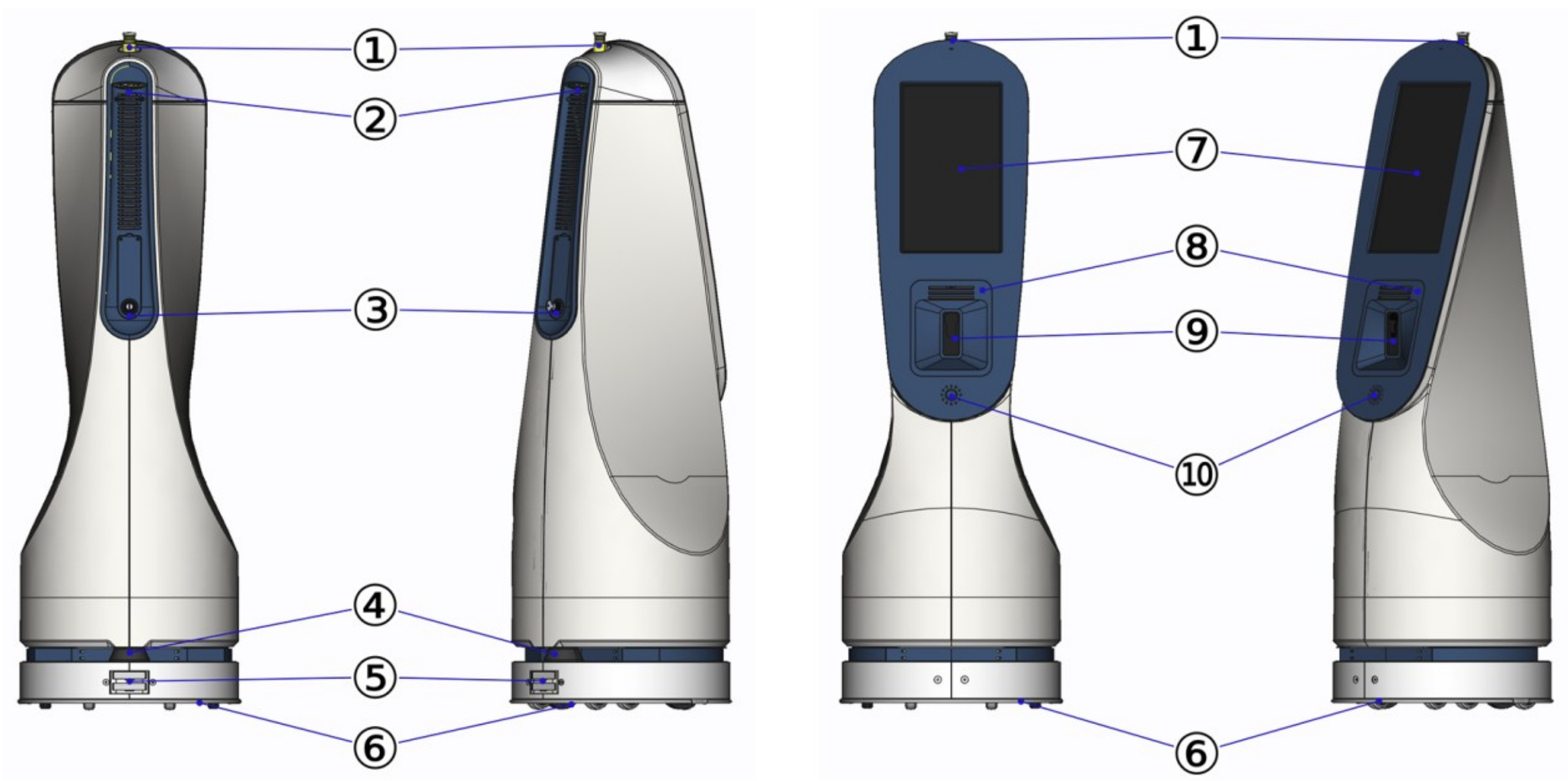

*Figure 1: The THERY platform. 1 - safety stop; 2 – obstacle avoidance camera; 3 – key lock; 4 – safety LIDAR; 5 – charging contacts; 6 – safety strip/bumber; 7 – touch display; 8 – reader for RFID transponder; 9 – depth camera (Azure Kinect); 10 – speaker.*

## 4.2 Software configuration

The on-board PC runs Ubuntu and MIRA, the robotic middleware (Einhorn et al., 2012), in which also parts of the applications are implemented. For navigational purposes, MetraLabs CogniDrive is used, which also allows the setting of no-go-areas.

### *4.2.1 Cloud-based therapy management system (TMS)*

THERY is administered by a cloud-based therapy management system intended to be used by staff of the healthcare facility only. It provides several levels of administration:

a) Administrative mode: user role management like admins, supervisors who can e.g. define certain training plans (type of training, training frequency per day, number of days per training), and set up new regular users, and regular users with access to mostly operative functions of the TMS.
b) Operating time mode: operating times of the robot can be flexibly defined, allowing e.g. to block robot use during busy times in the ward.
c) Setting up robot users: TMS users can set up robot users like patients or elderly by e.g. entering patient data. They can also assign training plans and certain robots to a dedicated robot user.
d) Training results: After training, TMS users can view the training results.
e) Printed training results can be exported into hospital information systems.

## 4.3 Functional modules of the robot

### *4.3.1 Free-run mode*

Sometimes hospital wards are crowded, and the robot may be in the way. Therefore, people around should be able to easily move the robot. However, once it is in operating mode, the motor controller reacts to any external force, making it hard to move the robot. For that purpose, a free-run mode was implemented via the emergency stop: once pressed, the motor control first halts the robot and then "unlocks" the robot, making it easy for bystanders to push it to a place where it does not disturb. This mode can also be used easily if the robot gets stuck and cannot free itself.

#### *4.3.2 Pain measurement*

Depending on the application, users are asked about their pain level. For this, they have a slider menu for the Visual Analog Scale (VAS). Depending on the outcome, the robot suggests not performing the training.

#### *4.3.3 Person re-identification*

The robot tracks the person it was assigned to during training. Once operating in an environment with multiple people around, it is possible that the robot could get confused and starts interacting with the wrong person. For that reason, the ROGER project used a person re-identification in form of a red card that a user had to wear, and the robot tracked that red card. This was a simpler implementation as in the ROREAS project (Gross, Scheidig, et al., 2017) where sensor fusion algorithms were used tracking the legs with a LIDAR and capturing full-body images of the person to track its full-body appearance. Instead, on THERY a face detector is implemented (Cognitec Systems GmbH/Dresden, Germany) that is also used for person identification at airports worldwide. In this context, the robot takes a picture of its user directly after log-in, and re-uses this image throughout the training procedure, before it is deleted at the end of the training.

#### *4.3.4 Administrative mode on the robot*

Admins can set waypoints for the training track in an administrative mode. Waypoints (e.g. for a patient signup position from which the robot moves to a start position to commute between that and a turning point position) can be selected by pushing the robot to these positions and store the localized position via screen buttons. The admin mode also allows manual (un-)docking with a charging station. For accessing the admin mode, a dedicated RFID transponder is necessary.

#### *4.3.5 Robot and exercise tutorials*

To allow for self-service on the robot, each application allows the user to access a general video tutorial on how to operate the robot. It explains how to handle mistakes of the robot. It encourages users to take a break, it reminds them to wear proper footwear to reduce fall risks, etc. Besides a general robot gait training tutorial, there are separate tutorials for each training program, for instance explaining the different walking modes in 2-point and 3-point gait.

#### *4.3.6 Patient training check*

Once users like patients log in via presenting a transponder, they must confirm some basic training aspects (for the crutch training: the side (left or right) on which they were operated), and the training program, and their user ID. This function mitigates patient risks associated with the training programs.

#### 4.3.7 Follow-mode

THERY instructs users to follow the robot. During the training, the robot keeps a constant distance to the user to obtain a full-body picture for generating a skeleton model for pose analysis. This means that the robot adjusts to the users' speed – up to its maximum speed of 0.8 m/s and subject to an obstacle-clear path (for details, see also (Vorndran et al., 2018)).

### 4.4 General training process

Fig. 2 outlines the general training process. It starts with using the TMS by defining a training plan and assigning an RFID transponder to the user (e.g. a patient). The robot THERY then receives a training plan and the associated transponder ID via WiFi. The user then can approach THERY by presenting the transponder, which's ID is read out, and loads the training plan. It provides introduction videos, takes an image for person re-identification and surveys the pain level (see both below). Then THERY drives to the training area where it commutes between two waypoints. During the training, person re-identification takes place, using a skeleton model generated with the image data from the Azure Kinect camera to identify the users' head, and the Cognitec software for face recognition. Subject to the training program, THERY may analyze the users' movements and output training feedback (see also the explanation of the THERY UAG application below). After a training session, THERY summarizes the walked distance and training time for the user, and transfers a more detailed report of the training via WiFi the TMS, where it can be viewed by the TMS users. The TMS is designed as a generic framework that allows implementing multiple applications executable on THERY.

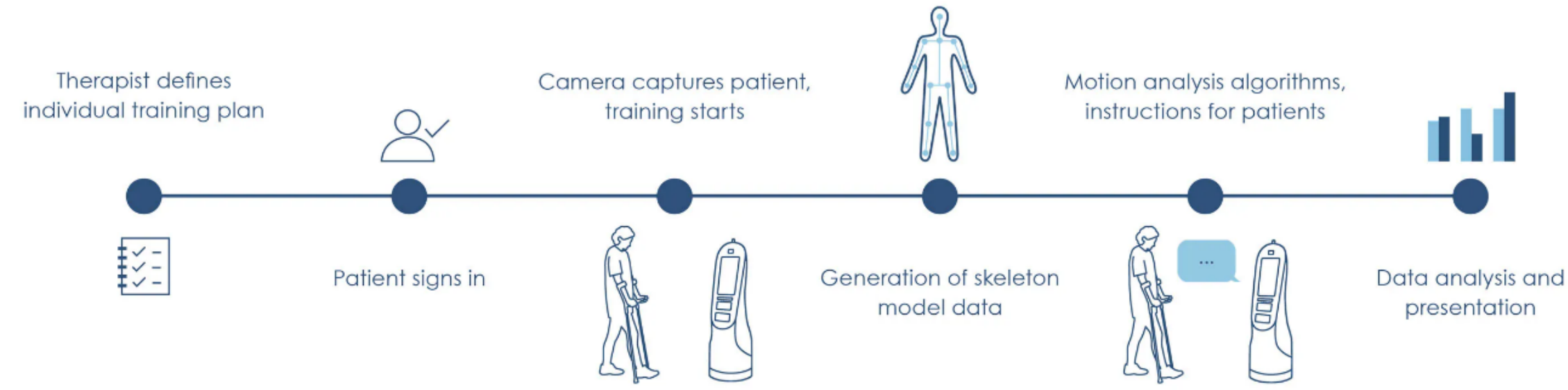

*Figure 2: General workflow for THERY-based trainings*

### 4.5 The THERY UAG application

Based on the workflow in Fig. 2, THERY UAG performs gait training on forearm crutches with patients after lower-limb surgery such as total hip or knee arthroplasty, ankle joint or leg fractures, etc. The motivation behind this application is to both assure that the patients use their crutches correctly and motivate them to exercise more to increase their mobility level, which helps them recover faster.

The robot can distinguish between the two common gait patterns known as 2-point and 3-point gait. In 3-point gait, the operated leg is placed between the two crutches (with basically 3 points – i.e. the two crutches and the operated leg – on a virtual line on the ground), mostly happening shortly after surgery, while in 2-point gait, a crutch and the contralateral leg are placed on one virtual line on the ground.

In this context, the robot uses the Azure Kinect camera (with 30 frames/second) and the Azure Kinect SDK for motion capture to generate a skeleton model. Moreover, the point cloud is analyzed and crutches are detected (see e.g. Scheidig et al. (2019)), for which subsequently the ground landing position of the crutches is evaluated jointly with the skeleton points.

Based on the skeleton points, feature classification allows the analysis of 18 different gait parameters, such as step length, step width, trunk inclination, and weight transfer. For them, deviations from desired movement patterns (i.e. "gait errors") are analyzed. Next, the detected gait errors are compared with a gait error prioritization triggered within a defined interval of several steps, and the application only outputs correction recommendations for the gait error with the highest prioritization (see e.g. Fig. 3). Within a subsequent step interval, the patient has time to correct the gait error, before THERY UAG searches for the next errors. This error prioritization had been developed in the ROGER project according to guidelines established by physiotherapy experts. After the training, all gait errors are transferred to the TMS, where physiotherapists can view the results.

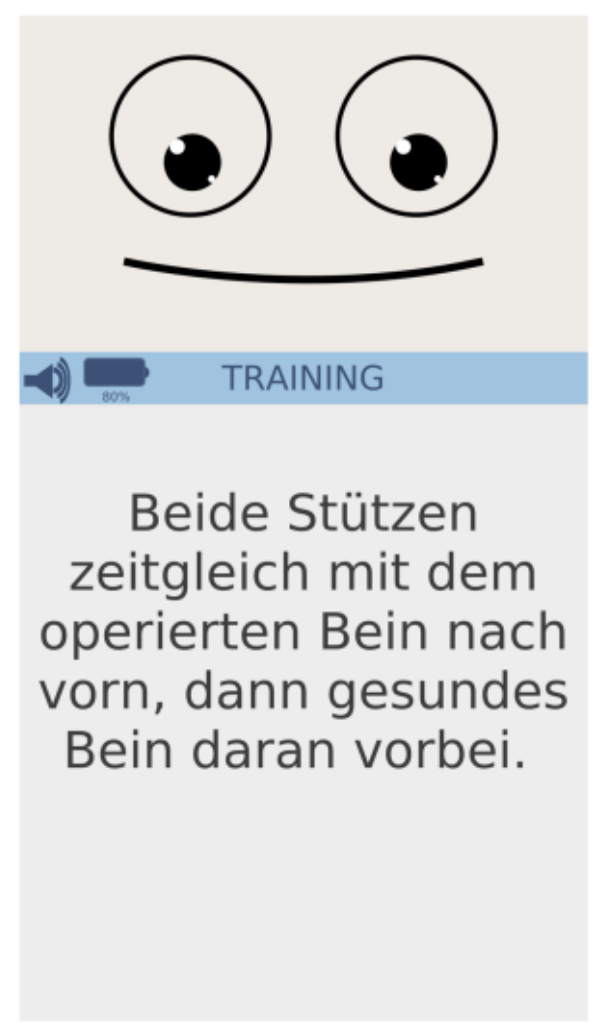

*Figure 3: Example for a training recommendation outputted within THERY UAG. There is an animated face in the upper part.*

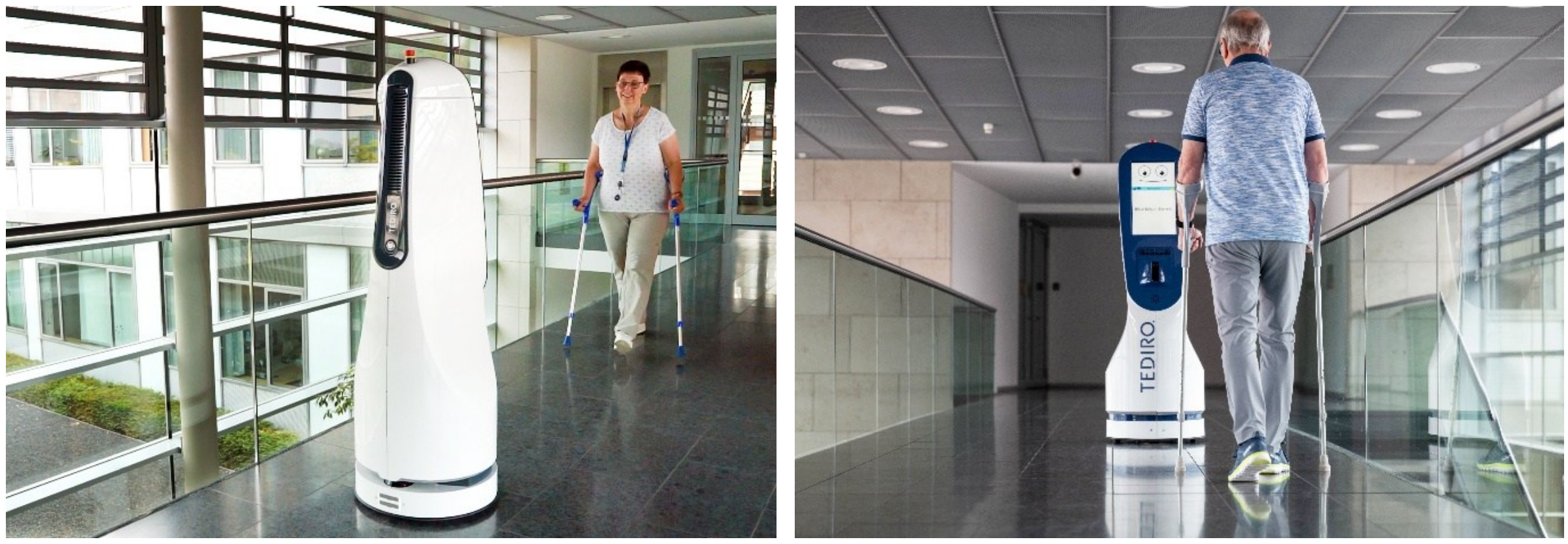

*Figure 4: Gait training on forearm supports with THERY UAG (consent obtained from all individuals shown)*

## 5 Requirements for walking training service robots in operational environments

As mentioned above, the robot THERY with its application THERY UAG and the cloud-based TMS have been certified as a medical device within the European Union by TEDIRO

Healthcare Robotics GmbH as manufacturer according to the EU Medical Device Regulation EUR 751/2017 (MDR). However, receiving such a certificate does not necessarily mean that the system is sufficiently complete for use in an operational environment (i.e. that it meets technology readiness levels higher than 6) as autonomous robots interacting with humans, in particular, face numerous technical challenges when interacting with their environment beyond their core functionality. MDR compliance, first and foremost, only means that the system meets its intended purpose/core functionality, is safe for patients and fulfills minimum requirements in terms of usability. It does not say anything about peripheral functions, which are pivotal for acceptance in practice and for reaching high technology readiness levels.

For that reason, a research project called TANGO was started that involved MetraLabs GmbH for some hardware components to elevate that robot to a higher technology readiness level, and develop two further applications on top of the robot platform THERY to increase the capacity utilization for use also within smaller facilities with low patient numbers.

## 5.1 THERY platform

For the THERY platform, MetraLabs worked on robot speed improvements and novel battery technologies.

### *5.1.1 Driving speed of the THERY platform*

The target robot speed of 1.4 m/s was selected based on published normative gait data indicating that healthy adults typically walk at speeds approaching 1.4 m/s under normal conditions (Bohannon and Williams Andrews, 2011; Brown et al., 2015). The challenge was that with higher driving speeds, the protection fields from the LIDAR increase, which possibly means that they reach a size at which multiple obstacles in the operative environment would fall into these fields, reducing the robots' speed to 0.3 m/s (the save speed) and making a proper training impossible.

Simulations of adjusting the breaking deceleration for reaching a safe speed of the platform after driving in "operative mode" revealed that the robot could potentially reach a driving speed of up to 1.5 m/s. With a deceleration of 1.5 m/s$^2$, THERY would reach the safe speed still with a small increase of the safety fields, which seems to be acceptable. Consequently, a new gearing unit was built into the platform that operates with the existing motors. Stability tests revealed that the robot passed two important regulatory hurdles.

a) The robot, indeed, could decelerate as simulated without being at risk of falling over.
b) There were also doubts beforehand regarding the step test of section 9.4.2.4.3 of the ISO 60601-1, requiring that it passes a 10 mm step when driving. Internally conducted tests showed that it did not fall over when passing a step of 10 mm during driving, and, importantly, it could pass the step in self-driving mode at maximum speed.

### *5.1.2 Alternative batteries*

The integrated mobile base of THERY is a modified TORY base, a robot from MetraLabs frequently used for shelf-scanning and manufactured at larger volume than the THERY platform. However, the batteries of the TORY platform only have a capacity of 40 Ah, which is not sufficient for THERY given the computing requirements for motion capture. Moreover, the LiFePO4 cell type has been in use for more than a decade now and is not state-of-the-art anymore. Recent improvements in battery cell technologies could be helpful to address these points.

Consequently, a new battery cell type was found that has a higher packaging density than the old cells. A new battery was designed that fits into the TORY space, with a capacity of 60 Ah. This means that there will be economies of scale for the THERY platform regarding the platform design.

## 5.2 Field tests for THERY UAG and subsequent implementations

Multiple field tests were performed that involved 8 acute care hospitals and 7 rehabilitation facilities. Some acute care hospitals also had an inpatient or large outpatient rehab unit. THERY was demonstrated to users like therapists, nurses, patients, and care home residents. They could test the robot and the TMS and were surveyed to obtain feedback for reshaping stakeholder requirements and adjust robot features. This was possible because the system was fully certified and fulfilled all requirements according to the EU Medical Device Regulation (i.e. had a CE-mark). The field tests did not only involve questioning the stakeholders, they also involved substantial logfile analysis to elicit causes of robot behaviors. The following obstacles were identified that posed adoption barriers.

### *5.2.1 Navigational challenges*

It was noticed that in some facilities where the tests took place, the robot abruptly stopped. Investigating the issue, it turned out that the obstacle avoidance camera (Orbbec Astra Stereo S U3) generated ghost points of obstacles in 3D space for some floor grounds. Consequently, the robot stopped to avoid potential collisions. A substantial improvement was achieved by integrating noise filtering capabilities as provided by an alternative version of the software driver module.

### *5.2.2 Corrective outputs*

Users of THERY UAG remarked that, there is little variation in the correction outputs when they make the same mistake repeatedly. They found this tiring and tedious. For each error category, two to three output variants were selected at random. In response to user feedback, the number of variants for identical gait errors was increased to provide greater variety.

### *5.2.3 Robot speed and training quality*

In particular, fitter patients who walk faster than the robot's maximum speed allows no longer benefit from gait training with THERY UAG. Due to their higher walking speed, the fitter individuals moved too close to the device, meaning that the user's face was no longer within the part of the frame that the camera could capture. This led to a loss of patient tracking and, consequently, an interruption to the training session. This challenge is to be addressed by increasing the robot's speed. To provide an interim solution in the meantime, a voice prompt has been created that instructs the user to maintain a sufficiently large distance from the robot. This indirectly results in the user following the robot more slowly and, consequently, walking more slowly.

### 5.2.4 *Auto-charging the robot*

To ensure the robot is available for training sessions throughout the day, it must be undocked from the charging station. Whilst the robot is in use, medical staff in the operating environment must monitor its charge level so that, should the charge become low or, at the latest, at the end of the agreed operating time, they can return the robot to the charging station. It was noted that this manual docking and undocking procedure relies too heavily on the availability of staff. In particular, at weekends and on public holidays, the robot should be able to offer training sessions regardless of whether staff are present. To resolve this issue, an automatic docking and undocking function has been added. This is triggered either by a low battery level or by the operating times configured on the robot.

### 5.2.5 *Training plan definitions and training results in the TMS*

Setting up training plans proved to be time-consuming for therapists, as each training session had to be scheduled manually. To reduce this effort, the planning functionality was redesigned to support the bulk creation of recurring training sessions. In addition, the system was extended to allow alternating training applications within a single training plan. Furthermore, a filtering feature for training results was implemented, so that therapists can easily select certain dates for viewing results in the dashboard.

## 5.3 Application development for THERY GO and subsequent field tests

### 5.3.1 *Requirements and application development*

Another goal of the project was the development of THERY GO, a gait training application designed for use both with various walking aids and without a walking aid.

For the initial implementation, stakeholder requirements were initially generated internally (given that TEDIRO employs people from the end user groups), and the existing THERY UAG application was modified in a first step. This included a deactivation of the walking pattern analysis and associated introductory videos. A different set of motivational outputs was defined, referring to the distance walked, the remaining training time of the person ("you already covered 50m – you are doing great") or the remaining distance ("just 2 minutes left in the training, well done so far"). The TMS was extended by a new application type for a THERY GO training THERY GO and THERY UAG training sessions can be mixed in a single training plan. Post-training survey questions have been adopted to the THERY GO application and consequently allow us to systematically collect user experience data.

### 5.3.2 *Field tests of THERY GO*

To enrich the stakeholder requirements, THERY GO was demonstrated in several field tests in acute care hospitals and rehab facilities, but also three nursing homes, i.e. actual operational environments.

Nursing home residents were able to use the robot without difficulties. Furthermore, hospitals indicated the potential of THERY GO for patients with obesity as a means of increasing motivation for physical activity. User statements obtained and implemented covered a more diverse set of motivational training feedback than initially implemented (similar like in section 5.2.2) and a user-configurable audio volume for the feedback outputs. Feedback from prospective operators and healthcare facilities also revealed the need for extended configuration options. Facilities should be able to define whether patients are required to complete an assessment of their pain level before and after a training session and which assessment instrument is to be used for this purpose, for example a pain scale or the Borg Scale. Furthermore, stakeholders expressed the requirement for more flexible handling of the introductory videos (tutorials) that were displayed on the robot prior to the start of a training session. Depending on the preferences of the respective facility, these tutorials should be disabled, made skippable by the user, or alternatively provided via existing bedside terminals. Additionally, a brief summary of the training results should be displayed at the end of each training session if desired by the facility. Parts of these features were also implemented for THERY UAG.

### 5.4 Application development for a walking speed test and subsequent field tests

#### *5.4.1 Requirements and application development*

Various standardized testing procedures are used in clinical practice to quantitatively assess different aspects of walking ability and mobility. Walking speed, for example, can be measured using the 10-meter walk test (10MWT), which measures the time taken to walk a distance of 10 meters. The 6-minute walk test (6MWT) is another commonly used assessment tool and is primarily used to evaluate functional walking ability and endurance based on the distance covered within six minutes.. Stakeholder interviews indicated that the 6MWT is used more frequently and requires more personnel resources, since it takes the most time to execute. Therefore, a decision was made to prioritize the implementation of the 6MWT on the THERY robot. However, apart from the limits of robot speed, it turned out that a robot moving ahead of a patient could eventually influence the patients’ walking speed. Consequently, an alternative solution was conceptualized in which the robot mainly serves as a stationary system, measuring the walking distance by monocular distance tracking. The concept included a workflow analysis and a user interaction concepts including outputs, which was implemented as a mockup in Figma. This mockup was discussed with stakeholders in a rehab facility leading to updated requirements. The required walking distance should be between 15 m and 30 m, preferring the latter since less systematic measurement errors are introduced by slowdowns at the turning points. Clear test instruction and progress indication to the patients are crucial for repeatable tests. Test abortion must be always possible and automatically documented. Discussions with stakeholders further revealed that assessment areas and training areas might be at different locations, and transitions between them must be handled by the system automatically.

Due to the time constraints (i.e. the project duration), implementation was not yet possible, meaning also that eventual adoption barriers of the actual system in rehab hospitals could not yet be identified.

## 6 Conclusion

While static robotic therapy systems are an established product category, mobile service robots in this space are in an early stage. For deploying such autonomous systems in clinical environments, it is necessary that they address more than their core clinical functionality to find acceptance. Examples encompass e.g. driving speed, for which certain regulatory constraints exist which must be managed. Another field was user interaction and the self-service mode, with variety in providing tutorials and motion feedback. Easily removing the robot when it gets stuck is another aspect. Auto-charging, known e.g. from cleaning robots, was another topic that is making the life of nurses easier when the robot operates outside of typical therapist working hours.

The field tests clearly showed that such complementary features going beyond the core clinical functions are pivotal for user acceptance. A further finding in this project was that the software applications in the field tests ran very stable. Colleagues who had developed service robots outside the healthcare field before attributed this to the IEC 62304 complaint development process with its detailed verification procedures, slowing down the development on the one hand, but reducing the need for fixing software bugs in the field later.

Overall, the project demonstrates that the development and commercialization of mobile service robots for healthcare applications require a long-term product stewardship approach. Sufficient resources for the continuous refinement and optimization of the system after product launch should already be considered during product planning to achieve a level of maturity that enables widespread acceptance and sustainable adoption in clinical practice.

## Acknowledgements

The authors would like to express their sincere gratitude to Alexander Vorndran, Thanh Quang Trinh, Alexander Leipnitz, André Laukner, Muhammad Abeer Akmal, and Andreas Adam for day-to-day feature implementation and support throughout the field tests of the work described in this publication. Parts of the work described herein were funded by the Free state of Thuringia, Ministry of Economy, Science, and Digital Society, co-funded by the European Union, grant number 1006177 (the TANGO project).